# Feature subset selection for Big Data via Chaotic Binary Differential Evolution under Apache Spark

Yelleti Vivek[1,2], Vadlamani Ravi[1*] and P. Radhakrishna[2]

[1]Center of Excellence in Analytics,
Institute for Development and Research in Banking Technology,
Castle Hills Road #1, Masab Tank, Hyderabad-500076, India

[2]Department of Computer Science and Engineering, National Institute of Technology,
Warangal-506004, India

yvivek@idrbt.ac.in; rav_padma@yahoo.com; prkrishna@nitw.ac.in

**Abstract**

Feature subset selection (FSS) using a wrapper approach is essentially a combinatorial optimization problem having two objective functions namely cardinality of the selected-feature-subset, which should be minimized and the corresponding area under the ROC curve (AUC) to be maximized. In this research study, we propose a novel multiplicative single objective function involving cardinality and AUC. The randomness involved in the Binary Differential Evolution (BDE) may yield less diverse solutions thereby getting trapped in local minima. Hence, we embed Logistic and Tent chaotic maps into the BDE and named it as Chaotic Binary Differential Evolution (CBDE). Designing a scalable solution to the FSS is critical when dealing with high-dimensional and voluminous datasets. Hence, we propose a scalable island (iS) based parallelization approach where the data is divided into multiple partitions/islands thereby the solution evolves individually and gets combined eventually in a migration strategy. The results empirically show that the proposed parallel Chaotic Binary Differential Evolution (P-CBDE-iS) is able to find the better quality feature subsets than the Parallel Bi-nary Differential Evolution (P-BDE-iS). Logistic Regression (LR) is used as a classifier owing to its simplicity and power. The speedup attained by the proposed parallel approach signifies the importance.

## 1. Introduction

Feature subset selection (FSS) [1,2] involves selecting a subset of features that is highly discriminative and informative. It is critical in the data pre-processing step with the following advantages: improves the model performance, avoids overfitting, improves the human comprehensibility and reduces the training time of the model. Majorly, FSS is solved in three different ways namely, Filter, Wrapper and Embedded approaches. Filter approaches involve the selection of the feature by using any statistical measures such as t-statistic/F-statistic, forward selection/backward elimination, mutual information (MI), information gain (IG) etc., whereas Wrapper methods involve formulating the problem as a combinatorial optimization problem, where a classifier or a regression model computes the fitness function for an evolutionary algorithm, which drives the entire process. Embedded approaches are those classification/regression techniques, where feature selection is performed by design, while the training takes place. Among the above-mentioned models: The complexity of the wrapper model is high and its performance is dependent on the underlying classifier as well as the optimizer has chosen. But the advantage is that the inter-relations between features are well-taken care, because of the combinatorial nature of the method. Many wrapper based EAs are proposed to solve FSS [4,14-16].

---

* Corresponding Author, Phone: +914023294310; FAX: +914023535157

Table 1: List of Notations

| *Notation* | *Meant for* |
|---|---|
| ps | Population Size |
| lps | Local / Sub- population size chosen for an island |
| N | Number of features |
| mMig | Maximum number of Migrations |
| mGen | Maximum number of Generations |
| MR | Mutation rate |
| CR | Crossover rate |
| Xtrain | Train Dataset RDD |
| Xtest | Test Dataset RDD |
| P | Population RDD |
| $cvalue_j$ | $j^{th}$ sequence number generated by chaotic map |
| $M_i$ | $i^{th}$ Mutation Vector |
| $T_i$ | $i^{th}$ Trail Vector |

Evolutionary Algorithms (EAs) do not always guarantee global optimal solutions. Chaos is a mechanism to handle the evolutionary dynamics and improves the adaptability of EAs, thereby influencing the decision making policy. Chaos is largely dependent on the initial conditions. Primarily, chaos has two properties: Ergodicity and Stochastically intrinsic nature. A chaotic map generates the sequence of numbers that exhibits the above chaotic properties. Hence, fusing EAs with a chaotic mechanism helps the algorithm to escape from the local minimum. It also facilitates to search in the regions which are left-out by the random sequence. Consequently, a chaotic map also adds diversity to the population thereby enhancing the quality of the solutions.

The volume of the datasets is increasing day by day in many fields. Sequential algorithms cannot handle such huge voluminous data. Hence, the prominence of distributed and parallel solutions have increased tremendously. Map Reduce [3] is a distributed programming paradigm designed with the sole purpose to serve as a helping hand in designing scalable and sophisticated solutions. The extant Feature subset selection (FSS) methods are not scalable to large voluminous datasets. On the other hand, parallel approaches are not yet proposed for this task. This motivated us to design a parallel FSS method that incorporates the chaotic principles in the evolutionary algorithm thereby enhancing their search capability in case of large voluminous datasets.

The main highlights of this work are as follows: (i) Proposing a novel single multiplicative objective function including both the AUC and the cardinality of the selected-feature-subset (ii) Incorporating the chaotic maps in the population initialization and crossover operator of BDE (resulting in CBDE) to attain population diversity and enhance the search mechanism (iii) Designing and developing parallel models based on island strategy (iS) under Apache Spark Environment to make the proposed models scalable.

The rest of the paper is organized in the following way: Section 2 presents literature review, while Section 3 presents the overview of the BDE and chaotic maps used in the current study. Section 4 presents the overview of the proposed approaches. Section 5 describes the proposed parallel island strategy for both the BDE and CBDE. Then, Section 6 gives the experimental setup, while Section 7 discusses the results. Finally, Section 8 concludes the paper.

## 2. Literature Review

Storn and price [4] proposed Differential Evolution (DE), a stochastic population based global optimization problem. Kohavi and John [5] are the pioneers in formulating the FSS as a combinatorial optimization problem and proposed wrapper based solutions. Several sequential wrapper based DE are already proposed. Mlakar et.al. [6] designed a wrapper based DE in the multi-objective environment (DEMO) and is applied for facial recognition systems. Ghosh et.al. [7] proposed a feature selection method for the hyperspectral image data based on the self-adaptive DE (SADE) wrapper method. Lopez et.al. [8] proposed permutation DE, which is of wrapper type where SVM as a classifier involves permutation based mutation and modified recombination operator with an aim to maintain diversity. Zhao et.al. [9] developed a two-stage method, where in the first stage critical features are selected by the fisher score and information gain. This is followed by the second stage where the modified DE is applied and SVM is incorporated as a classifier. Srikrishna et.al [10] proposed a quantum inspired DE (QDE) with LR as the classifier. They proved that QDE has attained more repeatability than the BDE. Krishna and Ravi [11] proposed Adaptive DE in a single objective environment by considering LR as the classifier. Li et.al [12] demonstrated that DE-SVM-FS is performing better in terms of accuracy than the DE and SVM alone. Wang et.al. [13] proposed a simultaneous way to do feature and instance selection based on DE-KNN.

Chaotic based models have occupied their importance in bringing diversity to the population. There are various feature selection strategies based on chaotic models exist. Arora et.al [14] proposed a novel chaotic feature selection on interior search algorithm (CISA) which outperformed ISA. Qasim et.al [15] proposed a chaotic binary black hole algorithm (CBBHA) to conduct feature selection on three chemical datasets. CBBHA uses chaotic maps to generate the control parameters only. CBBHA outperformed BBHA in terms of time complexity. Sayed et.al [16] uses the various chaotic maps to generate the control parameter in the crow search algorithm (CSA) and named it as CCSA. Senkerik et.al [17] proposed a chaotic based DE, jDE and SHADE and applied it to various benchmark problems to prove the importance of adapting the chaotic maps. Javidi and Hosseinpourfard [18] proposed a chaotic based genetic algorithm where they have used Tent Map in population initialization and Logistic map in Crossover and Mutation to generate the control parameters. They have applied them to the benchmark test functions. Snaselova and Zboril [19] proposed a genetic algorithm based on chaos theory where they have used Logistic map in the crossover operator which enhanced the convergence of the algorithm. The authors applied various test functions to prove its importance. Chuang et.al [31] proposed a chaotic variation on the binary particle swarm optimization (CBPSO) where the chaotic maps are embedded in the population initialization and inertia weight control parameter. CBPSO uses KNN as a classifier with leave-one-out-cross-validation (LOOCV). Assarzadeh et.al [32] proposed a chaotic based PSO with the inclusion of mutation operation. KNN is the underlying classifier to perform feature selection and applied on the medical datasets. Ajibade et.al[33] proposed a hybrid version of PSO and DE and employed chaotic principles in the population initialization and inertia weight adaption of PSO only. Ahmed et.al [34] proposed a chaotic multi-verse-optimizer (CMVO) based feature selection using KNN as the classifier. CMVO embeds the chaotic map in the inflation operator. Along with feature selection, chaotic DE models also impact various fields such as bakers yeast drying process [20], techno-economic analysis [21] etc.

Now the discussion continues with the distributed and parallel versions of DE. To handle high dimensional datasets and big datasets several parallel and distributed versions of EAs [22-23] have been proposed. There are various ways of implementing parallel DE under the Hadoop environment and the Zhou [24] conducted computational analysis on the population based and data based MR solutions. Teijeiro et.al. [25] designed parallel DE under Spark-AWS to solve benchmark optimization problems on the most popular parallel models: Master-Slave (MS) model and island (iS) model. It is found out that the MS model suffers from high communication overhead and the islands models are computationally feasible. Recently, Chou et.al. [26] designed a parallel DE which can be used to solve large scale clustering problems. Al-Sawwa and Ludwig [27] designed DE Classifier (SCDE) to handle the imbalanced data. SCDE works on euclidian distance and finds the first optimal centroid and assigns class accordingly. The major drawback in the existed sequential wrapper approaches is its unsuitability in applying for big datasets. Although they can be applied to big datasets the computational complexity involved is very high. Also as to the best of our knowledge, there is no parallel approach solving FSS. Due to its low complexity, we have chosen an island (iS) based scheme.

## 3. Overview of the Binary Differential Evolution (BDE) and chaotic maps

This section presents the background theory related to the *BDE* and the chaotic maps which are utilized in the *CBDE*. All the notations used in this paper are presented in Table 1.

### 3.1 Binary Differential Evolution

BDE is one of the most popular and robust stochastic population global optimization algorithms. It starts with randomly initializing the population which undergoes mutation and forms the mutant vectors. Thus formed mutant vectors underwent the crossover operation and forms the trail vectors. Selection operation is applied and replaces the parent solutions with their corresponding better trail vectors. Thus obtained better solutions are treated as the parent population for the next generation. This is continued till it reaches the maximum number (*MAXITR*) generations. BDE is too well known to be discussed in detail. It is to be noted that in the current study, *DE/best/1/bin* strategy is employed.

### 3.2 Chaotic Maps

The two very well-known chaotic maps adopted in the current study are listed here:

**Logistic map [35]:** This is one of the simplest chaotic maps. It is defined in the below Eq.(1).
$$d_{t+1} = lw * ( d_t * ( 1 - d_t )) \qquad (1)$$
where the constant lw [0,4] determines the behavior of this logistic map. If 3.56 < w < 4, it is observed to have chaotic behavior. In current research study, lw=4 is chosen.

**Tent Map [36]:** It exhibits the topologically conjugate behavior to the logistic map. It is defined as in the below Eq.(2).
$$d_{i+1} = \begin{cases} tw * d_i, & d_i < 0.5 \\ tw * ( 1 - d_i ), & d_i \geq 0.5 \end{cases} \qquad (2)$$
where the constant tw [0,2] determines the behavior of the tent map. If 1 < tw < 2, then the Tent Map observed to have chaotic behavior. In current study, tw=1.5 is chosen.

# 4. Proposed Methodology

## 4.1 Proposed Objective Function

When FSS posed as a combinatorial optimization problem then it will have two objective functions: (i) minimizing the cardinality of the selected-feature-subset (ii) maximizing the corresponding AUC. To quantify both of the above we proposed a novel single multiplicative objective function. The proposed objective function consists of the product of the normalized cardinality score and AUC achieved by it, given in Eq.(3). Maximizing this fitness score is the objective. The mathematical formulation of the objective function for the solution $p_i$ is represented as below:

$$\max(f(p_i)) = \max\left(AUC_{score_i} * \left(1 - \frac{\#SelectedFeatureSubset_i}{N}\right)\right) \quad (3)$$

where $AUC_{score_i}$ is obtained by regression (LR) evaluating the selected feature subset. N represents the maximum number of features present in the dataset and $\#SelectedFeatureSubset_i$ is the number of selected features / cardinality obtained by a solution vector $i$.

## 4.2 Proposed Chaotic Binary Differential Evolution (CBDE)

In this section, the proposed Chaotic Differential Evolution is discussed. Chaotic Differential Evolution also adopted the same *DE/best/1/bin* strategy. The block diagram of the CBDE is depicted in Fig.1.

**Chaotically initialize the population**

The CBDE starts with chaotically initializing the population and is treated as a parent population. It means the chaotic sequence which is generated by the chaotic maps are used to determine the probabilistic selection of features. Because of the chaotic properties, thus generated sequence ensures diversity in the population. Still, the initial seed value for the chaotic sequence is a randomly generated number where thereafter numbers in the chaotic sequence are generated by the underlying chaotic map. The same mechanism is employed throughout the algorithm wherever the chaotic map is used. After this, the Logistic regression (LR) evaluates the AUC and then the fitness score is obtained by the Eq.(3).

**Mutation:**

After this, the parent population undergoes the CBDE Mutation operation to generate the population of mutated solutions (M) is generated by the below Eq.(4).

$$M_i = s_b + MF * (s_1 - s_2) \quad (4)$$

where the $s_b$ is the best solution chosen from the population and $s_1\ and\ s_2$ are randomly chosen from the population, MF is between 0 and 1.

**Chaotic Crossover:**

Thus formed population of mutated solutions underwent the chaotic crossover operation and forms the population of trail vectors by following the Eq.(5).

$$t_{ij} = \begin{cases} m_{ij}, & if\ cvalue_j < CR\ and\ j \neq randi \\ p_{ij}, & if\ cvalue_j \geq CR\ and\ j \neq randi \end{cases} \quad (5)$$

where $t_{ij}, m_{ij}, p_{ij}$ represents the $j^{th}$ bit of the $i^{th}$ trail vector $T_i$, trail vector $M_i$, parent vector $P_i$ and $cvalue_j$ is the chaotic number produced by one of the chaotic map, CR is between 0 and 1.

**Selection:**

The corresponding feature subsets are obtained from the trail vectors $T_i$ and then trains the LR. Later the AUCs are computed and the corresponding fitness score is obtained by Eq.(3). Then the following selection operation is applied thereby replacing the worst parents with their corresponding children as mentioned in Eq.(6). The thus formed population is treated as the parent population for the next generation.

$$P_i = \begin{cases} p_i, & if\ f(P_i) > f(T_i) \\ t_i, & otherwise \end{cases} \quad (6)$$

As depicted in Fig.1, step B, C and D are repeated till the algorithm reaches its convergence criteria which is the maximum number of generations (*MAXITR*).

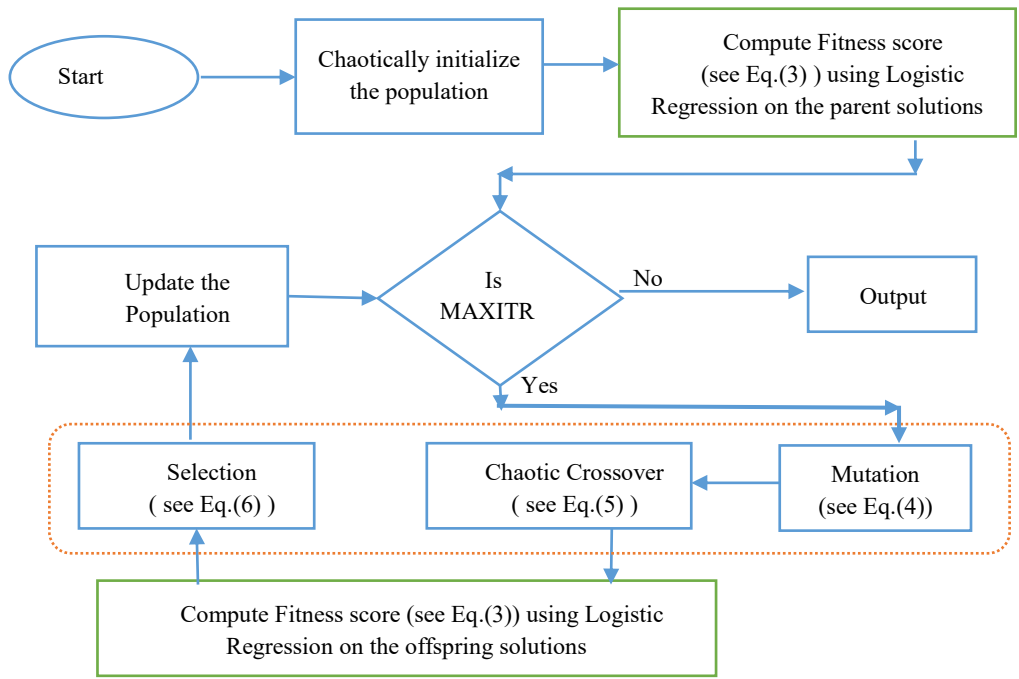

Fig 1: Block diagram of CBDE

## 5 Proposed Parallel Methodology

In this section, the proposed island model for both the BDE and CBDE are discussed. The common flow chart for both the P-BDE-iS and P-CBDE-iS is depicted in Fig.3.

### 5.1 Population structure

The population P follows the structure as depicted in Fig.2, where the *key* field is the solution unique identifier, The value field consists of the following sub-fields: the *Binary Encoded Solution* is a randomly initialized solution corresponding to the key, *Coef* are the coefficients obtained after training with the classifier, *Fitness score* is computed as per Eq.(3) and *AUC* field corresponds the computed AUC using the coefficients of the corresponding solution.

| Key | Value |
|---|---|
| $Key_1$ | $< Binary\ Encoded\ Solution_1, Coef_1, Fitness\ score_1, AUC_1 >$ |
| $Key_2$ | $< Binary\ Encoded\ Solution_2, Coef_2, Fitness\ score_2, AUC_2 >$ |
| …. | …. |
| $Key_{ps}$ | $< Binary\ Encoded\ Solution_{ps}, Coef_{ps}, Fitness\ score_{ps}, AUC_{ps} >$ |

Fig.2. Structure of the population

## 5.2 Parallel BDE island Model (P-BDE-iS)

The Driver algorithm which is executed on the Master node is presented in Algorithm 1 and the Worker algorithm which is executed on the Worker nodes is presented in Algorithm 2. The initial population is randomly generated and the hyperparameters, namely, *CR, MF, lps, mMig, mGen* are initialized by the Driver algorithm of *P-BDE-iS*. All these initialized population and hyperparameters are broadcasted to worker nodes. In the island approach, the *Xtrain* is partitioned into a few partitions/data islands where the number of islands is dependent on the size and nature of the dataset. Now, by calling the BDE mapper function, the control is shifted from the driver to the worker nodes thereby executing Algorithm 2. Each data island uses a randomly chosen subset of size *lps (where lps < ps)* from the initialized population and is referred to as local population / sub-population.

Thereafter, the train-and-update phase (see Algorithm 2) begins, where, by using the *Binary Encoded solution* column of each solution, the reduced dataset is obtained and the corresponding AUC is evaluated by the LR model. The fitness score is computed by using Eq.(3). The trained coefficients are given by the LR, fitness score, AUC are updated in the corresponding *Fitness score, Coef, AUC* field respectively. This is performed on the whole local population.

The *Binary Encoded Vector* is extracted from the local population upon which the Crossover and the Mutation operations are applied resulting in the formation of the trail vectors. The trail vectors are stored in the *local offspring population*, which also follows the structure as depicted in Fig.2. Then, the train-and-update phase is called on this *local offspring population.* Thereafter, the selection operation is performed, where the better child solution replaces the worst parent solutions based on the fitness score. The above procedure is repeated for the *mGen* number of generations. All the above operations are performed on an island. This ensures that the evolution of the local population on an island is done in parallel.

After completion of all the BDE mappers, the individually evolved sub-population are collected at the driver. Now the control is shifted to the driver. Then the *migration-rule* is applied where thus collected local population from all the islands are sorted as per their fitness score and the top *ps* solutions are only considered as the population for the next migration. The above procedure is repeated for *mMig* iterations. After the completion of *mMig* iterations, the coefficients are obtained from the evolved population. Using this information AUC is evaluated on the *Xtest*.

Algorithm 1: P-BDE-iS Driver Algorithm

*Input: ps, lps, P, Xtrain, Xtest, mMig, mGen, MF, CR, n, N*
*Output: P : population evolved after mMig migrations*

*1: i ← 0;*
*2: P ← Randomly Initialise the population*
*3: Divide the X into k number of islands*
*4: whille ( i < mMig){*
*5:     Call BDEislandMapper (namely, Algorithm 2)*
*6:     **Migration Rule:***
*       a) Total population ← Collect all the evolved populations from all the islands*
*       b) Sort Total population in the descending order of the fitness scores*
*       c) P ← Consider top ps solutions in sorted Total population*
*7:     i=i+1 }*
*8: coef ← collect coefficients of all the solutions*
*9: Evaluate Test AUCs using coef of the evolved population P*
*10: return P, Test AUCs*

Algorithm 2: P-BDE-iS Worker Algorithm

*Input: lps, mGen, MF,CR, N*
*Output: P : population evolved after mG, max generations*
*1: function BDEiSlandMapper{*
*2:     k ← 0;*
*3:     localP ← Randomly pick the lps size of subset from P         // local population*
*4:     **train-and-update phase:***
*       a) Evaluate LR on localP.BinaryEncodedVector*
*       b) Store thus computed coefficients and compute fitness score*
*          & update localP*
*5:     whille ( k < mGen){*
*6:         localPVector ← localP.BinaryEncodedVector*
*7:         mutatedVector ← perform Mutation on localPVector*
*8:         TrailVector ← perform Crossover on mutatedVector*
*9:         Create localOffspringP using TrailVectors           // local offspring population*
*10:        Perform **train-and-update phase** on localOffspringP*
*11:        localP ← Perform selection by combining localP and localOffspringP*
*12:        k=k+1 }*
*13: return localP}*

Algorithm 3: P-CBDE-iS Driver Algorithm

> **Input:** ps, lps, P, Xtrain, Xtest, mMig, mGen, MF, CR, n, N
> **Output:** P : population evolved after mMig migrations
>
> 1: i ← 0;
> 2: P ← Chaotically Initialise the population
> 3: Divide the X into k number of islands
> 4: whille ( i < mMig){
> 5:     Call CBDEislandMapper ( namely Algorithm 4)
> 6:     **Migration Rule:**
>         a) Total population ← Collect all the evolved population from all islands
>         b) Sort Total population descending order of their fitness scores
>         c) P ← Consider top ps solutions in sorted Total population
> 7:     i=i+1 }
> 8: coef ← collect coefficients of all the solutions
> 9: Evaluate Test AUCs using coef of the evolved population P
> 10: return P, Test AUCs

Algorithm 4: P-CBDE-iS Worker Algorithm

> **Input:** lps, mGen, MF,CR, N
> **Output:** P : population evolved after mG, max generations
> 1: function CBDEiSlandMapper{
> 2:     k ← 0;
> 3:     localP ← Randomly pick the lps size of subset from P  // local population
> 4:     **train-and-update phase:**
>         a) Evaluate LR on localP.BinaryEncodedVector
>         b) Store thus computed coefficients and compute fitness score
>            & update localP
> 5:     whille ( k < mGen){
> 6:         localPVector ← localP.BinaryEncodedVector
> 7:         mutatedVector ← perform Mutation on localPVector
> 8:         TrailVector ← perform Chaotic Crossover on mutatedVector
> 9:         Create localOffspringP using TrailVectors // local offspring population
> 10:        Perform **train-and-update phase** on localOffspringP
> 11:        localP ← Perform selection by combining localP and localOffspringP;
> 12:        k=k+1 }
> 13: return localP}

## 5.3 Parallel Chaotic-BDE Data island Model (P-CBDE-iS)

The P-CBDE-iS also follows the same parallelization strategy and the same is reflected in the Driver algorithm given in Algorithm 3 and the Worker algorithm is given in Algorithm 4. The major difference between the *P-BDE-iS* and *P-CBDE-iS* is the way the population initialization is done and the chaotic operation. This says that the train-and-update phase, as well as migration rule, are same for both *P-BDE-iS* and *P-CBDE-iS*. As the underlying parallel approach is identical for both the algorithms we have used a common flowchart, is depicted in the Fig.3. In the current study, logistic and tent map are studied. The logistic map variant of the *P-CBDE-iS* is named as *P-CBDE-iS-LM*. It uses logistic map in both the chaotic initialization step as well as in the chaotic crossover operation. Similarly, Tent Map used in both the chaotic initialization step as well as in the chaotic crossover operation is named *P-CBDE-iS-TM*.

P-CBDE-iS starts the driver algorithm in Algorithm 3 by chaotically initializing the population which follows the structure as depicted in Fig.2. After broadcasting the necessary information, a CBDE mapper (see Algorithm 4) is called on the data islands. This shifts the control from the master to the worker nodes. Similarly, as in *P-BDE-iS,* a *local population* of size *lps* is randomly chosen from the population. The *train-and-update phase* is performed on the *local population.* Thereafter the *Binary Encoded Vector* is extracted from the *local population* upon which the Chaotic Crossover and the Mutation operations are applied results in generating the trail vectors. Using this trail vectors information the *local offspring population* is formed upon which the train-and-update function is performed. This is repeated for the *mGen* number of generations. Then the migration rule is applied and forms the evolved population. The above procedure is repeated for *mMig* iterations. After the completion of *mMig* iterations, the coefficients are obtained from the evolved population. Using this information AUC is evaluated on the *Xtest*.

## 5.4 Classifier Algorithm

In the current study, Logistic Regression (LR) classifier is chosen because of: low complexity, quick training time, absence of hyperparameters and its non-parametric. Owing to its simplicity and power LR is chosen as a classifier.

Table 2: Description of the Datasets

| Name of the Dataset | #Datapoints | #Features | Size of the dataset |
|---|---|---|---|
| Epsilon [28] | 5,00,000 | 2000 | 10.8 GB |
| IEEE Malware [29] | 15,50,000 | 1000 | 3.2 GB |
| OVA_Uterus [30] | 1584 | 10935 | 108.3 MB |

## 6 Experimental Setup

This section describes the configuration of the cluster environment. All the experiments are carried out in the with cluster configuration: Total Number of nodes in the cluster: 05 (Master Node + 04 Slave Nodes). All the nodes including the master node have the Intel i5 8[th] generation with 32GB RAM each.

## 7 Results & Discussion

The description of the benchmark binary classification datasets is presented in Table 2. We performed stratified random sampling on each of the datasets in Table 2 and split them into train and test set ratio of 80%-20%. All the proposed approaches are executed under the same environment and the Hyper-parameters are as follows: for Epsilon dataset, ps=50, CR=0.9, MF=0.2, for IEEE malware dataset, ps=50, CR=0.91, MF=0.2 and OVA_Uterus dataset, ps=300, CR=0.95 and MF=0.005. All the parallel models are conducted for 20 experiment runs.

### 7.1 Three-way comparative analysis

The comparative analysis between the *P-BDE-iS* and *P-CBDE-iS* about the AUC and cardinality are given in Table 4. The cardinality of the feature subset selected by *P-CBDE-iS* is very less when compared to the *P-BDE-iS* in all the datasets, yet maintaining the similar or better AUC. The higher average cardinality shows that the solutions identified by *P-BDE-iS* are redundant when compared to the solutions given by *P-CBDE-iS*. This is more prevalent, especially in Epsilon and IEEE Malware datasets. This happened mainly because of the introduction of the chaotic maps in the population initialization step which helped in attaining more diversity, thereby enhancing the search mechanism nature of the algorithm. Chaotic Crossover makes sure that the algorithm is looking at the left-out search spaces, thereby reducing the redundancy of the selected feature subset. Due to this *P-CBDE-iS* model solutions stands better than the *P-BDE-iS*.

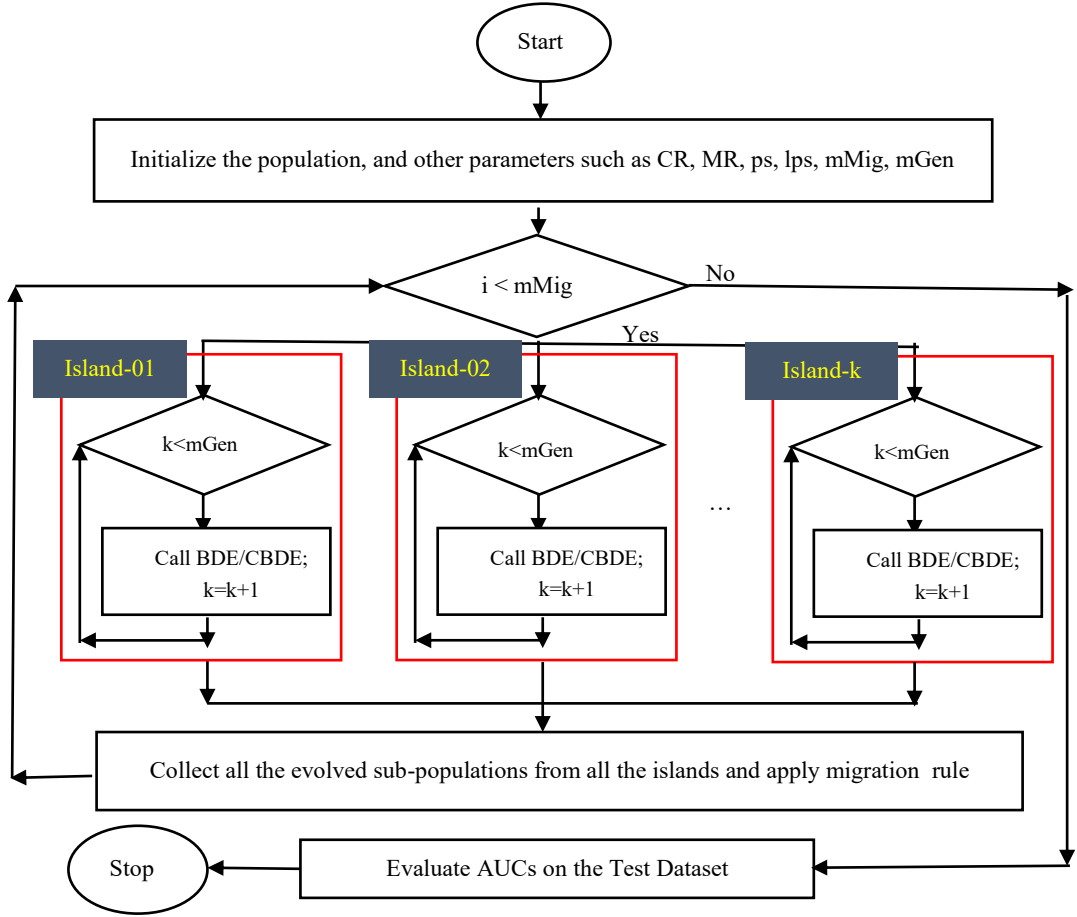

Fig 3: Flow chart for both P-BDE-iS and P-CBDE-iS based wrapper

Table 3: Average AUC and Cardinality achieved by all models

| Dataset | P-BDE-iS | | P-CBDE-iS-LM | | P-CBDE-iS-TM | |
|---|---|---|---|---|---|---|
| | Avg. Cardinality | Mean AUC | Avg. Cardinality | Mean AUC | Avg. Cardinality | Mean AUC |
| Epsilon | 405.45 | 0.797 | **328.8** | **0.801** | 251.35 | 0.779 |
| IEEE Malware | 338.95 | 0.796 | **277.55** | **0.797** | 221.1 | 0.776 |
| OVA_Uterus | 86.25 | 0.892 | **85.2** | **0.905** | 63.8 | 0.875 |

Even though the solutions given by the *P-CBDE-iS-TM* model are achieving less AUC than the *P-BDE-iS-LM,* it has to be noted that the cardinality of the selected feature subset is also very less than the *P-BDE-iS* and *P-CBDE-iS-TM*. Owing to this fact it can be said that *P-CBDE-iS-TM* is suffering from a slower convergence rate. This shows that the *P-CBDE-iS-TM* may achieve better solutions if allowed for more generations. But, doing so will increase the computational complexity.

## 7.2 Repeatability

Repeatability is one of the most important factors which embodiments the robustness of the model. The more repeatability the better the robustness of the model. In this section, the top most repeated solution which is repeated for at least 40% is considered with the cardinality and the AUC is presented here. #s represents the cardinality of the top-most feature subset with it, AUC. Results are accommodated in Table 4. *P-CBDE-iS-LM* achieved a better AUC with far less cardinality of the selected feature subset when compared to the *P-BDE-iS*. Whereas *P-CBDE-iS-TM* achieved very less cardinal feature subset, the AUC is also dipped. Hence in terms of repeatability with a better feature subset *P-CBDE-iS-LM* stands as a clear winner when compared to the rest of the two models. The top solutions given by all the models are having this repeatability shows the robustness of the algorithm.

Table 4: Most repeated feature subset selected by all models

| Dataset | P-BDE-iS | | P-CBDE-iS-LM | | P-CBDE-iS-TM | |
|---|---|---|---|---|---|---|
| | #s | AUC | #s | AUC | #s | AUC |
| Epsilon | 393 | 0.798 | **322** | **0.803** | 272 | 0.780 |
| IEEE Malware | 342 | 0.797 | **263** | **0.798** | 241 | 0.777 |
| OVA_Uterus | 86 | 0.893 | **85** | **0.901** | 61 | 0.877 |

## 7.3 Speedup

Speedup is defined as the ratio between the time taken by the sequential algorithm and the time taken by the parallel algorithm. It is represented as below:

$$\text{Speed Up (S.U)} = \frac{\text{Time taken by Sequential Version}}{\text{Time taken by Parallel Version}}$$

Table 5: Speedup of the proposed parallel models

| Dataset | Sequential E.T | P-BDE-iS | | P-CBDE-iS-LM | | P-CBDE-iS-TM | |
|---|---|---|---|---|---|---|---|
| | | E.T | S.U | E.T | S.U | E.T | S.U |
| Epsilon | 14707.06 | 4729.22 | 3.10 | 4619.54 | 3.18 | 4527.89 | **3.24** |
| IEEE Malware | 13145.23 | 5635.71 | 2.33 | 5418.84 | 2.42 | 5078.56 | **2.59** |
| OVA_Uterus | 3952.72 | 2043.42 | 1.93 | 2031.56 | 1.94 | 1896.56 | **2.08** |

Speedup obtained by the parallel algorithm is phased out in Table 5. All the parallel models are evaluated for the same number of migrations and generations hence the execution time (E.T) is almost similar for all the models. As four worker nodes are used in our experimental setup the maximum speedup that can be achieved is 4. The island models have achieved S.U from 1.93 to 3.24. The results evident that the island (iS) model has achieved S.U.

## 7.4 t-test analysis

In this section, the two-tailed t-test analysis is conducted on the fitness scores attained by *P-CBDE-iS-LM*, *P-CBDE-iS-TM* and *P-BDE-iS* over the 20 experimental runs. The significance level is taken as 5%. The degree of freedom is 20+20-2=38. The null hypothesis is *H0: both the algorithms are statistically equal, H1: both the algorithms are statistically different.* All the p-values as tabulated in Table 6, shows that the *P-CBDE* and its two variants are statistically different from the existed *P-BDE*. Hence it is derived that the chaotic and random sequences are different. This test also derives that the chaotic sequences generated by both the Logistic and Tent Maps are also quite different from each other. With this, it is proved *P-CBDE-iS-LM* and *P-CBDE-iS-TM* are statistically significant than the *P-BDE-iS* model.

Table 6: t-test analysis conducted between all models

| Dataset | P-BDE-iS vs P-CBDE-iS-LM | | P-BDE-iS-LM vs P-CBDE-iS-TM | | P-CBDE-iS-LM vs P-CBDE-iS-TM | |
|---|---|---|---|---|---|---|
| | t-statistic | p-value | t-statistic | p-value | t-statistic | p-value |
| Epsilon | 15.63 | 3.8e-18 | 9.16 | 3.7e-11 | 18.28 | 2.0e-20 |
| IEEE Malware | 9.44 | 1.61e-11 | 3.50 | 0.001 | 10.22 | 1.81e-12 |
| OVA_Uterus | 5.14 | 7.82e-6 | 10.28 | 6.4e-13 | 15.23 | 1.08e-18 |

## 8 Conclusion

The *P-CBDE-iS* model outperformed *P-BDE-iS* in terms of selecting better quality solutions. Both the models *P-CBDE-iS* with its two variants and *P-BDE-iS* shows the repeatability property. The speedup evident that the proposed parallel strategy is scalable. The T-test analysis shows that the *P-CBDE-iS-LM* with Logistic is quite significant in exploration and exploitation than the *P-BDE-iS* and *P-CBDE-iS-TM*. The current versions are non-adaptive. Hence in future making these models adaptable by using a self-adaptive mechanism and carrying out the same problem but in bi-objective and tri-objective environments. Few other chaotic map performances on the feature selection can also be investigated. From the parallelization strategy point of view: the current parallel version is designed with synchronization junctions, designing asynchronous parallel strategy will also be investigated.